\pdfoutput=1

\documentclass[11pt]{article}

\usepackage[final]{acl}

\usepackage{times}
\usepackage{latexsym}

\usepackage[T1]{fontenc}

\usepackage[utf8]{inputenc}

\usepackage{microtype}

\usepackage{subfiles}

\usepackage{inconsolata}
\usepackage{xspace}

\usepackage{tipa}

\usepackage{graphicx}
\usepackage{booktabs}
\usepackage[inline]{enumitem}
\usepackage{ulem}
\usepackage{accents}

\usepackage{color, colortbl}
\usepackage{todonotes}

\usepackage{tcolorbox}

\definecolor{mygreen}{rgb}{0.3, 0.73, 0.09}
\definecolor{myred}{rgb}{0.82, 0.1, 0.26}

\definecolor{zeroshot_color}{RGB}{009, 102, 240}
\definecolor{cot_color}{RGB}{017,214,255}
\definecolor{fewshot_color}{RGB}{074,255,185}
\definecolor{nickel_color}{RGB}{186, 252, 070}

\newcommand{\method}{\textsc{LingoLLM}\xspace}

\definecolor{myhighlight}{rgb}{0.804,0.98,0.77}

\title{\textit{Hire a Linguist!}: Learning Endangered Languages\\ in LLMs with In-Context Linguistic Descriptions}

\author{Kexun Zhang$^1$\quad\quad\quad
  Yee Man Choi$^1$\quad\quad\quad
  Zhenqiao Song$^1$ \\
  \textbf{
  Taiqi He$^1$ \quad\quad\quad
  William Yang Wang$^2$\quad\quad\quad
  Lei Li$^1$}\\
  $^1$Carnegie Mellon University \quad \quad
  $^2$UC Santa Barbara\\
  \texttt{\{kexunz,yeemanc,zhenqias,taiqih,leili\}@cs.cmu.edu} \quad \quad
  \texttt{william@ucsb.edu}
}

\begin{document}
\maketitle
\begin{abstract}
How can large language models (LLMs) process and translate endangered languages?
Many languages lack a large corpus to train a decent LLM; therefore existing LLMs rarely perform well in unseen, endangered languages. 
On the contrary, we observe that 2000 endangered languages, though without a large corpus, have a grammar book or a dictionary. 
We propose \method, a training-free 
approach to enable an LLM to process unseen languages that hardly occur in its pre-training.
Our key insight is to demonstrate linguistic knowledge of an unseen language in an LLM's prompt, including a dictionary, a grammar book, and morphologically analyzed input text.
We implement \method on top of two models, GPT-4 and Mixtral, and evaluate their performance on 5 tasks across $8$ endangered or low-resource languages. Our results show that \method elevates translation capability from GPT-4's $0$ to $10.5$ BLEU for $10$ language directions.
Our findings demonstrate the tremendous value of linguistic knowledge in the age of LLMs for endangered languages.
Our data, code, and model generations can be found at \url{https://github.com/LeiLiLab/LingoLLM}.

\end{abstract}

\section{Introduction}

Large language models (LLMs) are already powerful in many language understanding and generation tasks~\cite{brown2020language,ouyang2022training}.
Their language processing capabilities rely on very large amounts of training data \cite{kaplan2020scaling,hoffmann2022training}.
For example, a recent LLM Llama-2 uses a pre-training dataset with $2$ trillion tokens~\citep{touvron2023llama}.
While languages such as English or Spanish enjoy abundant accessible data, the majority of the world's 7000 languages lack a rich corpus,
including most endangered languages recognized by UNESCO \citep{moseley2010atlas}.
Existing LLMs such as Llama~\cite{touvron2023llama} and GPT-4 show poor performance on languages that may not occur in pre-training \cite{robinson2023chatgpt}.
The previous single machine translation model LegoMT~\cite{yuan-etal-2023-lego} supports 433 languages, the largest number of languages, but it cannot extend to endangered languages. 
We believe that speakers of endangered languages deserve equitable access to NLP technologies including LLMs. 
How can we enable an LLM with language processing capabilities on unseen and endangered languages?

\begin{figure}[t]
  \centering
  \includegraphics[width=0.49\textwidth]{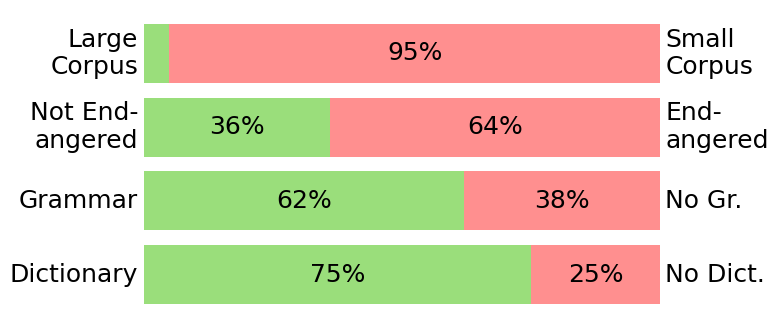}
  \caption{Among the world's $\sim$7000 languages, 95\% don't have enough data (>100K sentences) for training LLMs \cite{bapna2022building}, while most have a grammar book (60\%) or dictionary (75\%) \cite{nordhoff2011glottolog}, including many endangered languages \cite{moseley2010atlas}.
  Therefore, we utilize these linguistic descriptions to bring LLMs to endangered languages.}
  \label{fig:resource_comparison}
\vspace{-10pt}
\end{figure}

We are motivated by how human linguists analyze utterances in a language they don't know --- they use existing grammar books and dictionaries.
Fortunately, thanks to the efforts of generations of linguists over the years, many endangered languages have published dictionaries and descriptive grammar.
Compared to LLMs' training corpora, which mostly consist of unstructured text, these linguistic descriptions have two major differences.
First, they are instructional. Though they are much smaller than typical training sets, they contain explicit grammar rules of a language that can be used as instructions for both LLMs and humans.
Second, linguistic descriptions have much broader coverage. As shown in \autoref{fig:resource_comparison}, very few languages have training corpora, but most have documented grammar or dictionary.
However, directly using these linguistic descriptions in an LLM's prompt is infeasible.
A grammar book and a dictionary are often too large to fill in the prompt of an LLM.

\begin{figure}[th!]
  \centering
  \includegraphics[width=0.47\textwidth]{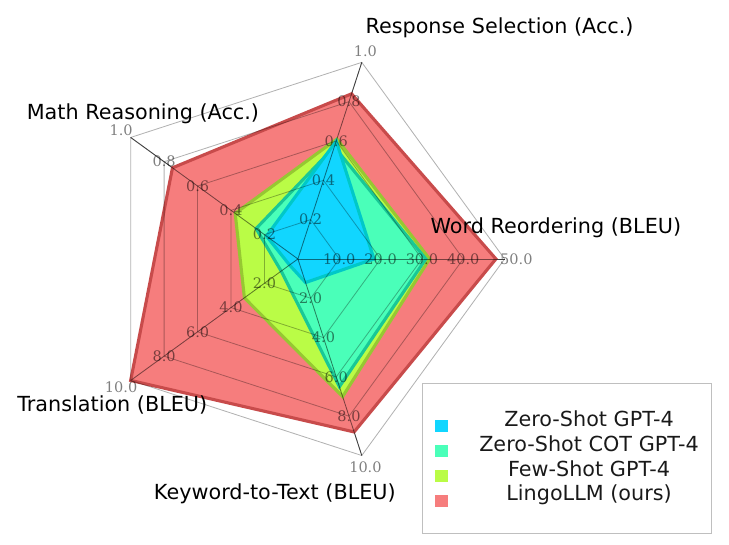}
  \caption{\method significantly outperforms GPT-4 on 5 NLP tasks across 8 endangered or low-resource languages.}
  \label{fig:performance}
\vspace{-10pt}
\end{figure}

In this paper, we propose \method, an efficient approach to enable an LLM to process and translate unseen languages that never occur in its pre-training.
Our key insight is to properly exploit linguistic description of an unseen language, including a dictionary, a grammar book, and morphologically analyzed input text.
\method first preprocesses input text in an endangered language  via a morphological analyzer and a dictionary, both from linguistic descriptions of the language.
The inputs, annotated with grammar features and word-level translations, are passed to an LLM along with the grammar book.
The LLM then translates the endangered language inputs to a high-resource language like English to process them.
\method is training-free as it only requires the underlying LLM to be instruction-tuned.
\method can adapt to languages according to the availability of different types of linguistic descriptions.

We implement \method on top of two models, GPT-4 and Mixtral. 
Our experiments consist of a total of 5 tasks (including translation from/to English, mathematical reasoning, response selection, word reordering, and keyword-to-text) in 8 endangered/low-resource languages that are geographically and typologically diverse.
As shown in \autoref{fig:performance}, \method significantly improves GPT-4's performance on all 5 tasks by a large margin. Noticeably the translation quality increases from an incomprehensible 0.5 to 10 BLEU points.

Our contributions are:
\begin{itemize}[topsep=0pt,leftmargin=*, itemsep=-2.5pt]
    \item We propose \method, an approach to integrate linguistic descriptions to process and translate text in endangered languages.
    \item With the help of linguists, we build processing systems for 8 typologically and geographically diverse endangered or low-resource languages according to the availability of different linguistic descriptions.
    \item Our experiments show superior performance of \method on all tasks, compared to strong baselines (GPT-4 and Mixtral). \method elevates translation capability from GPT-4's $0$ BLEU to $10.5$ BLEU for $10$ language directions. It improves GPT-4's mathematical reasoning accuracy from 18\% to 75\%, and response selection accuracy from 43\% to 63\%.
\end{itemize}

\section{Related Work}

Various recent studies explore the possibility of LLMs for low-resource languages on machine translation \cite{hendy2023good} and other NLP tasks \cite{ahuja2023mega, huang2023not}.
Their scope of evaluation mostly covers languages whose resources are low but still exist. %
Hence LLMs can still have a non-zero translation ability.
We go beyond their scope towards languages that are truly extinct where LLMs' zero-shot translation of them is near \textit{zero}.
Moreover, our method relies on external linguistic descriptions rather than internal knowledge of LLMs, focusing on how LLMs can utilize information they don't know instead of how they can ``recall'' information they have seen in training.

\noindent \textbf{Evaluating LLMs for Low-Resource Languages.} 
Many \cite{jiao2023chatgpt, hendy2023good, zhu2024multilingual} suggest that LLMs do poorly on low-resource languages. \citet{robinson2023chatgpt} evaluated ChatGPT's machine translation performance on 204 languages and found that ChatGPT consistently underperforms traditional machine translation models on low-resource languages.
\citet{ahuja2023mega} evaluated several LLMs on 16 NLP tasks and found significant performance drops on low-resource languages.
While their conclusions corroborate our motivation, the languages they evaluate are very likely to exist in LLMs' training set, as indicated by LLMs' significantly positive performance in the zero-shot setting.
On the contrary, the focus of our paper is on endangered languages that are truly extinct in the training data with near-zero performance from LLMs.

 \begin{figure*}[th!]
  \centering
  \includegraphics[width=0.95\textwidth]{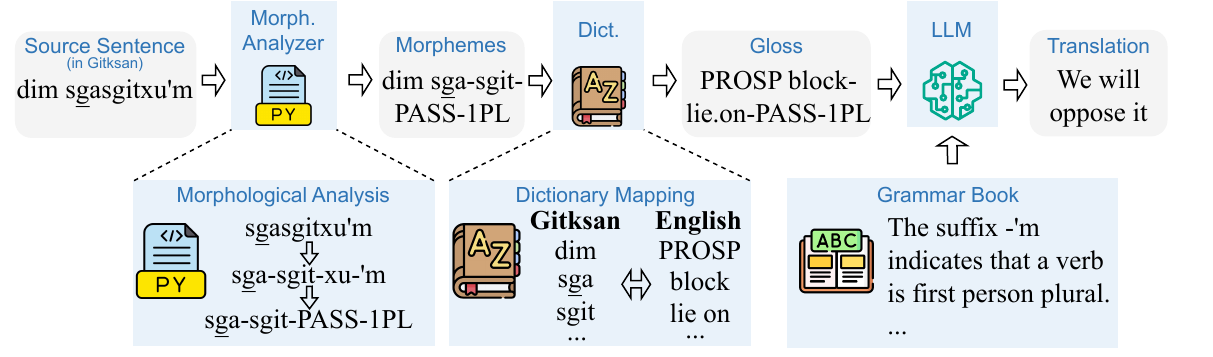}
  \caption{\method uses a morphological analyzer to transform the source sentence into morphemes, looks up the morphemes in a dictionary to obtain the gloss, and finally feeds both the gloss and a grammar book to an LLM to obtain the result.}
  \label{fig:pipeline}
\vspace{-10pt}
\end{figure*}

\noindent \textbf{Improving LLMs for Low-Resource Languages.}
Existing studies improve LLMs' performance on low-resource languages via prompt engineering.
Few-shot prompting \cite{ahuja2023mega} puts input-output pairs in the prompt as exemplars using different selection strategies.
Chain-of-thought prompting further prompts the model to solve the problem step by step, either by explicit instruction \cite{huang2023not} or by in-context examples of step-by-step answers \cite{shi2022language}.

Another line of work relies on external modules and knowledge.
\citet{ahuja2023mega} uses commercial translation systems to first translate a task to English before giving it to LLMs.
\citet{gao2023design} augments the input sequence with POS tagging.
However, both commercial translators and POS taggers are hard to find for the languages we evaluate.
Our paper is closest to \citet{tanzer2023benchmark}, where they also depend on dictionaries and grammar books to translate an endangered language with LLMs.
Their main goal was to propose an uncontaminated translation benchmark for evaluation purposes on a single endangered language.
Compared to \citet{tanzer2023benchmark}, we go beyond machine translation to multiple NLP tasks and evaluate the proposed \method extensively on 8 endangered/low-resource languages that are diverse.
Also, we make use of morphological analyzers, an extra symbolic module that can be derived from grammar books.

\section{The \method Approach}

Since we do not have enough data to fine-tune an LLM, \method equips an LLM with linguistic knowledge to process text in an endangered language (\autoref{fig:pipeline}).
We obtain a morphological analyzer, a dictionary, and a grammar description for each target language. 
Our method consists of four steps: 
\begin{enumerate*}[label=\arabic*)]
    \item Given an utterance in an endangered language, we first use a morphological analyzer to split each word into \textit{morphemes};
    \item We search for the closest matches from a dictionary for each morpheme to obtain an annotated \textit{gloss};
    \item We prompt an instruction-tuned LLM with the annotated gloss and a grammar description to get the \textit{translation} in a high-resource language such as English; and
    \item We further process the translated text using a same LLM for the downstream task. 
\end{enumerate*}

To adapt \method to a new language, we first collect linguistic descriptions with the help from linguists.
We use three types of linguistic descriptions in \method~-- morphological analyzers, dictionaries, and grammar books.
For all languages studied in this paper, we look for references to these descriptions on Glottolog\footnote{\url{https://glottolog.org/}} and then collect them via web interfaces or ebooks.
For one type of description, we build a universal interface to use them despite different underlying formats.
We list the collected linguistic descriptions in \autoref{app:lind}.

\subsection{Morphological Analysis: \\ \quad\quad\  Source Sentence $\rightarrow$ Morphemes}

A morphological analyzer is a program that maps a word to a sequence of morphemes, the smallest meaningful constituents.
Morphemes include \textit{stems} that indicate concrete meanings and \textit{linguistic features} that indicate grammatical roles.
For example, an English morphological analyzer might map \textit{cats} to \textit{cat +Noun +Plural}, where \textit{cat} is a stem and \textit{+Noun} and \textit{+Plural} are two features.
Features make it easier for people and LLMs to find out the grammatical roles of a word, while stems are more convenient for dictionary search than their inflections and derivations.

We use existing finite-state morphological analyzers to identify the stem and features of each word in the source sentence.
These analyzers are written as finite-state transducer using the python implementation of foma \cite{hulden-2009-foma}.
We directly apply them to the source words to obtain the morphemes.
For example, in \autoref{fig:pipeline}, the Gitksan word \textit{s\underline{g}asgitxu'm} is transformed into four morphemes \textit{s\underline{g}a-sgi-PASS-1PL}, where the stems are \textit{s\underline{g}a} (meaning ``to block'') and \textit{sgi} (meaning ``to lie on''), and the features indicate that it's a verb with a passive voice whose subject is first person plural.

One concern that may be raised about using morphological analyzers is their availability because they may not exist for as many languages as grammar books do.
While there are no statistics about the availability of these analyzers, this issue can be resolved by having trained linguists create  morphological analyzers from grammar books.

\subsection{Dictionary Mapping:\\
\quad\quad\  Morphemes $\rightarrow$ Gloss}

We use a dictionary to map the stems (morphemes with concrete meanings) to their dictionary definitions.
The word-level translations, along with the stems, constitute the gloss of the source sentence.
The gloss can then be utilized by the LLM to formulate sentence-level translation.
The dictionary mapping process is not as straightforward as it seems and can involve multiple steps.

\noindent \textbf{Step 0. Normalizing the script.}
Before we implement the mapping from source words to their translations, we must make sure the scripts used by the dictionary and the input are the same as those in the test set.
This is often not the case, especially for endangered languages.
For example, the Manchu dictionary we use \citep{norman2020comprehensive} represents the phonemes /\textipa{tS\super{h}}/, /\textipa{S}/, /\textipa{u}/ as q, x, and v, while our Manchu inputs represent the three phonemes as c, š, and ū.
We manually compare the written forms of the same words in the input and the dictionary to derive rules that map one script to another.

\noindent \textbf{Step 1. Deciding the input: words or stems?}
We can either use source words or their stems as the input to the dictionary, depending on the availability of a morphological analyzer and the scope of words in the dictionary.
Usually, it is easier to find matches in the dictionary for word stems produced by a morphological analyzer.
But we have to use the original words when such an analyzer is not available.
Many dictionaries only have entries for one form of a verb.
For example, Manchu dictionaries might only contain verbs in their present tense form with the suffix \textit{-mbi}.
When this is the case, we have to get the stems of the words first or use some sort of fuzzy matching.

\noindent \textbf{Step 2. Finding the closest match.}
Online dictionaries' search algorithms often provide multiple possible matches for a word. In the case where we are unable to retrieve the word stem or the word stem does not exist in the dictionary, we would not be able to find an exact match from the dictionary and need to choose the closest match using the edit distance. 
For instance, the word stem for the Gitksan word \textit{mismaaxwsxum} (a plural marking attribute meaning white) is \textit{mismaaxwsxw}, but we are unable to find \textit{mismaaxwsxw} within the Gitksan dictionary we use \citep{gitksan_mothertongues}. However, we can find the following partial match, \textit{maaxwsxw} or \textit{maxwsxw} meaning ``to be white'', \textit{maaxws} meaning ``snow (on ground)'', \textit{misaax} meaning ``daylight'', and \textit{sawnsxw} meaning ``paper''. Using the edit distance as a selection metric, we can retrieve the closest matches \textit{maaxwsxw} or \textit{maxwsxw} that are most related to the word \textit{mismaaxwsxum}.

\noindent \textbf{Step 3. Collecting other relevant words.}
Some dictionaries' entries contain references to other entries.
The content of these referenced entries provides complementary information related to the matched word.
For example, the entry for \textit{qoohiyan} in our Manchu dictionary states that it stands for ``Korea'' the place. It also references another Manchu word \textit{solho}, meaning ``Korean'' the people.
To collect such information, we traverse the graph formed by cross-entry links starting from the match until all connected entries are found or the number of found entries exceeds a threshold.

\subsection{Incorporating Grammar Knowledge\\
\quad\quad\ Gloss $\rightarrow$ Translation and Beyond}

Lots of word-level grammatical information is already covered in the morphemes produced by morphological analyzers.
However, some very important information, such as what the subject of the sentence is or what noun is an adjective modifying, can still be unknown.
Therefore, we prompt the language model with grammar knowledge to give further guidance.

We obtain such knowledge of grammar from grammar books of different languages.
For books that are scanned, we use optical character recognition (OCR) to transform them into pure text.
If the size of the book fits the context length of a language model, we directly put the entire book in the prompt.
Otherwise, we use GPT-4 to generate a summary of the grammar which is able to fit in the prompt.
Once the translation of the source sentence is created, the LLM can then follow the instructions and process the sentence as required.

\section{Experiment}

\begin{table*}[th]
\small
\centering
\begin{tabular}{lcccccc|cccc|c}
\toprule
                   & \textbf{mnc} & \textbf{git} & \textbf{usp}& \textbf{ntu}& \textbf{ddo}& \textbf{wol}& \multicolumn{2}{c}{\textbf{arp}}& \multicolumn{2}{c}{\textbf{bzd}}& \textbf{Avg.} \\ 
                   & $\rightarrow$en   & $\rightarrow$en& $\rightarrow$es& $\rightarrow$en& $\rightarrow$en& $\rightarrow$en& $\rightarrow$en& en$\rightarrow$& $\rightarrow$es& es$\rightarrow$& \\ \midrule 
& \multicolumn{10}{c}{\textbf{GPT-4}} & \\
Zero-Shot          & 0    & 0 &  0.1& 0 & 0& 3.9& 0& 0.2& 0.4& 0 &0.5\\
Zero-Shot CoT      & 0.7   & 0  &  0.3& 0 & 0& 11.4& 0.4& 4.1& 0.4& 0.1&2.4\\
Few-Shot           & 0.5  & 9.3 &  2.2& 0 & 0.8& \textbf{13.5}& 1.0& 2.2& 0.8& 1.7&3.2\\
\rowcolor{myhighlight} \method {\scriptsize\ dict. only}& 8.3   & 7.7 &  10.7& 11.7 & 11.1& 6.9& 6.0& 14.5& 2.7& 2.2&8.2\\
\rowcolor{myhighlight} \method  & \textbf{10.8}  & \textbf{14.3} &  \textbf{12.4}& \textbf{12.9} & \textbf{15.1}& 8.1 & \textbf{9.4}& \textbf{15.6}& \textbf{4.3}& \textbf{3.0}&\textbf{10.5}\\ \midrule

& \multicolumn{10}{c}{\textbf{Mixtral-8x7B}} & \\ 
Zero-Shot          & 0.2  & 2.0 &  0.3& 1.2& 0.8& 7.4& 0.8& 0.5& 0.2& 0& 1.3\\
Zero-Shot CoT      & 0.5 & 3.4 &   0.2& 1.3& 0.4& 6.2& 0& 0.7& 0.5& 0.1& 1.3\\
Few-Shot           & 0.5 & 4.0 &  2.2& 2.2& 0.6& 8.6& 0.9& 0.5& 1.7& 1.8& 2.3\\
\rowcolor{myhighlight} \method {\scriptsize\ dict. only}& 4.1  & 4.7 &  3.9& 6.3& 6.0& 6.0& 5.2& 7.3& 2.6& 1.3& 4.7\\
\rowcolor{myhighlight} \method  
                   & \textbf{4.4}  & \textbf{7.9} & \textbf{4.6}& \textbf{7.3}& \textbf{10.7}& 3.2&  \textbf{7.4}& \textbf{8.4}& \textbf{3.0}& \textbf{2.2}&\textbf{5.9} \\

\bottomrule
\end{tabular}
\caption{\method significantly improves LLMs' ability to translate between low-resource/endangered languages and high-resource ones (such as English and Spanish).
The zero-shot performance of GPT-4 and Mixtral on these languages is near zero for 7 out of the 8 languages measured by spBLEU.
\method increases the BLEU score to 10.5 on average for GPT-4.
The languages are labeled using their ISO 639-3 code.
See \autoref{app:code}.
}
\label{tab:translate_spbleu}
\end{table*}

\begin{table*}[th]
\small
\centering
\begin{tabular}{p{1.5cm}p{4.1cm}p{4.4cm}p{4.2cm}}
\toprule
                & \centering{\textbf{Manchu}} & 
                \centering{\textbf{Gitksan}} & \multicolumn{1}{c}{\textbf{Arapaho}} \\ \midrule
 Input& suweni geren xusai dorgi de nikan i niyalma udu qoohiyan i niyalma udu& Way ts'ax wildiihl hehl Gitwinhlguu'l ii needii hasakdiit ehl reserve. "Needii hasaga'm dim dip suwii gi'namhl laxyibi'm," dihiida.&nihcihcee3ciiteit niiyou nuh'uuno heenees3i'okuutooni'\\ \midrule
Few-Shot        &  Every person in the military and every person in the common people must have courage      &  He said, "I will stay here in Gitanyow, and you will go to the reserve. 'You will learn to speak English well there,' he told me."       &    I'm going to work for you tomorrow.     \\ \midrule
\method          &  How many Chinese people and how many Koreans are there among your numerous students?      &  "Although it seems that the people of Kitwancool don't want the reserve, 'We do not wish to give away our land,’” they said.       &      Someone accidentally entered this room where people sit.   \\ \midrule
Ground Truth    &  Among your many students, how many are Chinese and how many are Korean?      & And now even though the people of Kitwancool said they did not want the little reserve; "We don't want to give away our land," they said.        &     He inadvertently walked in where peope were sititng .    \\ \bottomrule
\end{tabular}
\caption{Example translations produced by \method, compared to ground truth translation and the few-shot baseline.
Note that the translations from few-shot prompting are nonsensical and completely irrelevant to the actual translation.
More examples in \autoref{tab:translate_ex_app}.}
\label{tab:translate_ex}
\end{table*}

\subsection{Experiment Setup}

The benchmark data, code, and model generations can be found in the supplementary material.
The prompts we used are listed in \autoref{sec:app_prompts}.
We ran all of our experiments on two LLMs - GPT-4's checkpoint \texttt{gpt-4-1106-preview} and the open-weights model Mixtral-8x7B.
Note that we run Mixtral with 4-bit quantization.
We sample 1 output for each input at the sampling temperature of 0.8.

\subsubsection{Baselines}

\textbf{Zero-shot prompting.}
We directly prompt the model with text in the low-resource language and instruction in English.
The model is informed of the source language and the type of task to perform.

\noindent \textbf{Few-shot prompting.}
We randomly sample 3 examples from the validation set of the data as in-context demonstrations.
We use the exact same examples for all data samples.
The prompt only contains the input and output of the examples.

\noindent \textbf{Zero-shot Chain-of-Thought.}
We prompt the model with instructions like ``solve this problem step by step''.

\subsubsection{Benchmarks and Metrics}

\textbf{Translation.} For Manchu (mnc), we manually collect 70 parallel sentences from \textit{Nogeoldae}, a textbook of colloquial Chinese and Manchu published in 1705 containing various dialogs in both languages. We manually translate the Chinese sentences to English.
For Gitksan (git), Natugu (ntu), Arapaho (arp), Uspanteko (usp), Tsez (ddo), we use the parallel corpus provided by \citet{ginn-etal-2023-findings}, as well as their provided gloss.
We randomly sample 100 sentences from the corpora for each of these languages.
For Bribri (bzd), we use data from AmericasNLP 2023 Shared Task \cite{ebrahimi-etal-2023-findings}.
For Wolof (wol), we use data from Flores-200 \cite{nllb2022}.
We evaluate using spBLEU \cite{goyal-etal-2022-flores}, with the SentencePiece tokenizer of Flores-200.

\noindent \textbf{Conversation Understanding.}
To evaluate whether \method can improve LLMs' understanding of discourse in endangered languages, we construct a response selection benchmark automatically.
We collect passages or conversations in Manchu, Gitksan and Arapaho, and extract context-response pairs from these conversations.
For each context, we sample 3 other irrelevant responses.
The model is given a context and 4 responses and tasked to select the correct one.
Model performance is evaluated by the number of contexts for which the model can select the correct response.
To avoid the known order bias of LLMs \cite{zheng2023judging}, we shuffle the order of the choices for each context-response pair and average the accuracy.

\noindent \textbf{Math Reasoning.}
We evaluate how \method can solve reasoning tasks in endangered languages with mathematical problems.
Following \citet{shi2022language}, we collect their Chinese translation of GSM8K \citep{cobbe2021training} problems and hire native speakers of Manchu to translate them into Manchu\footnote{They were paid beyond the local minimum wage.}.
They are instructed to filter out problems with concepts that are infrequent in Manchu and replace the units with the ones that are more common.
After sampling and filtering, we obtain 20 math word problems.
We evaluate the performance by the number of problems solved.

\noindent \textbf{Word Reordering and Keyword-to-Text.}
To evaluate whether \method can learn the sentence structure of endangered languages, we evaluate it on two tasks -- sentence reordering and keyword-to-text.
We evaluate sentence reordering in three languages -- Manchu, Gitksan, and Arapaho.
We take 70 sentences in each language and randomly shuffle the word order in each sentence.
The shuffled sentence is then given to the language model to find the correct order.
Keyword-to-text is a more difficult task, where we manually select content words from each sentence, shuffle their order, and give them back to LLMs to create sentences based on these keywords.
Since this task is annotation-expensive, we only evaluate it in Manchu with 30 sentences.
We measure the quality of both tasks using spBLEU.

\subsection{Results}

\textbf{Translation.}
\method enables translation for endangered languages. We report \method's performance on translation in \autoref{tab:translate_spbleu}.
We only include the translation direction for the language if the corresponding linguistic descriptions are easily accessible. 
On 9 out of 10 translation directions, \method can significantly improve the LLMs' performance.
For GPT-4, the average increase in spBLEU is 10.5.
For Mixtral, the average increase is 5.9.
The Bribri translation from and to English exhibits the least improvement in BLEU, which is largely due to the low dictionary coverage.
Both zero-shot baselines have BLEU smaller than~1 for most languages except for Wolof to English and Arapaho from English directions, indicating that LLMs have very little knowledge about these endangered languages. 
Among the languages, various baselines for Wolof to English translation demonstrate good results.
Since the English parallel of Wolof came from Wikipedia and is included in the Flores dataset, the high performance of these baselines is susceptible to potential contamination \cite{robinson2023chatgpt}.
Few-shot prompting is the best baseline for all three languages.
We hypothesize that this could be because few-shot demonstrations and test data for these endangered languages might come from the same book.
Even so, the translations from few-shot prompting are still mostly irrelevant, as demonstrated by the examples in \autoref{tab:translate_ex}, where the few-shot translations are completely off the topic.

\begin{figure}[t]
  \centering
  \includegraphics[width=0.48\textwidth]{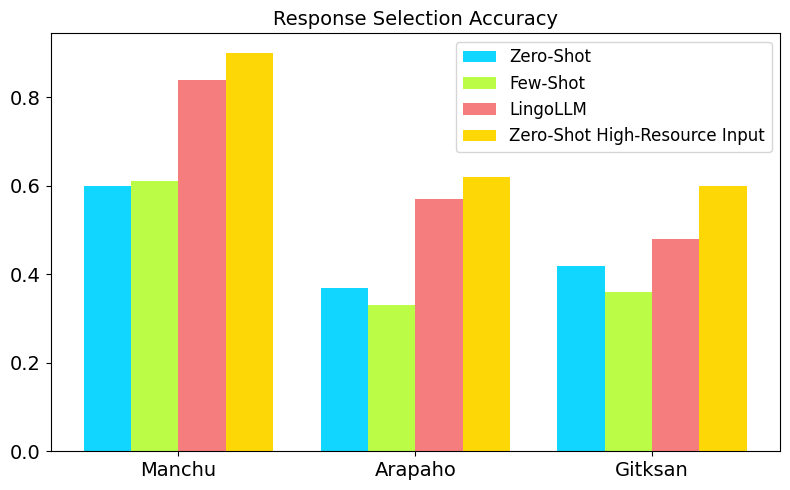}
  \caption{\method significantly improves LLMs' ability to select correct responses. On all three endangered languages, it achieves a performance comparable to high-resource language inputs.}
  \label{fig:understanding}
\vspace{-10pt}
\end{figure}

\noindent \textbf{Response Selection.}
Other than the baselines, we also compare \method with the zero-shot inputs in high-resource language.
Since the original conversations are written in parallel with high-resource language input, zero-shot with the high-resource language inputs is considered the upper bound of the performance.
As demonstrated in \autoref{fig:understanding}, \method improves GPT-4's response selection accuracy for all three languages, with 20\% improvement for Manchu and Aarapaho and 6\% improvement for Gitksan.
Note that for Manchu and Arapaho \method's performance is only 6\% lower than the supposed upperbound.
This indicates that \method significantly improves LLMs' ability to understand discourse in an endangered language.
Note that it is not surprising that zero-shot GPT-4 has a 70\% performance because word overlaps can be a decent indicator of correct responses.

\noindent \textbf{Mathematical Reasoning.}
As demonstrated in \autoref{fig:manchu_only_tasks}, \method can significantly improve the mathematical reasoning ability of LLMs on Manchu.
Zero-shot and few-shot baselines do not exceed 40\% accuracy while \method solves 75\% of the problems.
One surprising finding is that unlike translation where GPT-4's performance is near zero,
it actually has a positive zero-shot performance on math reasoning.
It might be due to contamination as the original problems in English are from GSM-8k, a widely distributed dataset.
On the other hand, the superior performance of \method corroborates its translation ability, because a very precise translation of a math question is needed for the model to answer it correctly.
We found that when questions involve concepts that are less common in endangered languages, it's easier for \method to fail.

\noindent \textbf{Word Reorder and Keyword-to-Text.}
As demonstrated in \autoref{fig:manchu_only_tasks}, \method improves GPT-4's performance on word reordering and keyword-to-text.
Compared to zero-shot, \method is 8x better on keyword-to-text and 2.5x better on word reordering.
These improvements indicate that LLMs equipped with \method are able to generate more coherent sentences in endangered languages.

\begin{table}[t!]
\small
\centering
\begin{tabular}{lccc}
\toprule
                & \textbf{Math} & 
                \textbf{Keyword} & \textbf{Word} \\ 
                & \textbf{Reasoning} & \textbf{to Text} & \textbf{Reorder}\\ \midrule
 Zero-Shot  & 18.7\%& 1.2& 18.4\\
CoT & 25.0\%& 7.0& 31.0\\
Few-Shot        &     37.5\%& 6.5&      31.8\\

\method          & \textbf{75.0\%}&  \textbf{8.8}&  \textbf{47.9}\\ \midrule
High-Res& 100\%& N/A& N/A\\
\bottomrule
\end{tabular}
\caption{ On math reasoning, keyword-to-text and word reordering, \method significantly improves GPT-4's performance.}
\label{fig:manchu_only_tasks}
\end{table}

\begin{table*}[th]
\small
\centering
\begin{tabular}{p{2cm}p{2cm}p{3.5cm}p{6cm}}
\toprule
                 & BLEURT & Input Example                               & Translation Example                                 \\ \midrule
Dict. Only        & 0.4573 & what shoot goose moose                      & Did the {\color{myred}\underline{moose}} shoot the grouse?                     \\
Dict. + Morph. Analysis & 0.5448 \textbf{(+19\%)} & what-CN CONTR shoot-TR-\textbf{2SG} grouse IRR moose & What did {\color{mygreen}\underline{you}} shoot over there, a grouse or a moose? \\ \midrule
Ground Truth     & -      & -                                           & What did {\color{mygreen}\underline{you}} shoot? A grouse or a moose?  \\ \bottomrule         
\end{tabular}
\caption{Grammar features produced by morphological analysis significantly improves \method's performance by 19\%.
As the example demonstrates, the feature \textbf{2SG} indicating ``second person singular'' helps the model to identify the correct subject of the sentence -- ``you'', while the stem-only baseline has the wrong subject -- ``moose''.}
\label{tab:morph_abl}
\end{table*}

\section{Ablation and Analysis}
\label{sec:analysis}
We conduct ablation studies and qualitative analysis to show how helpful each component of \method is and explore the best ways in which they can be used.
Note that some of our ablation experiments depend on extra annotations such as oracle dictionary mappings and oracle glosses, which only exist for a subset of the languages.

\subsection{Morphological analysis helps.}
To examine whether morphological analysis can provide extra information for \method, we analyze the results of the Gitksan translation test set with and without morphological analysis.

As shown in \autoref{tab:morph_abl}, morphological analysis significantly improves BLEURT score by 19\%.
The example in the table demonstrates that with morphological features, important information such as the number of nouns and the tense of verbs can make a huge difference in translation quality.

\subsection{High-quality dictionary helps.}

\begin{figure}[h!]
  \centering
  \includegraphics[width=0.45\textwidth]{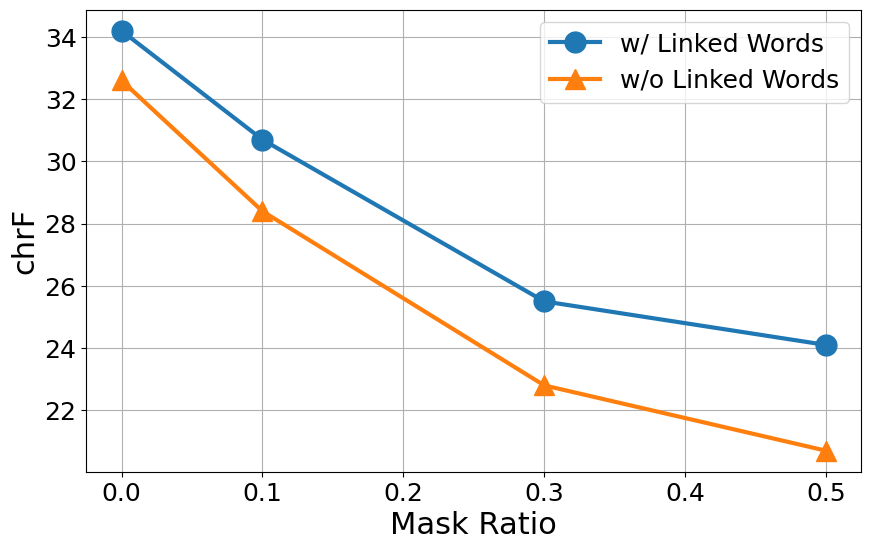}
  \caption{When more dictionary entries are masked out, \method's performance drops. When other relevant words referenced in the dictionary are not considered, \method's performance also drops.}
  \label{fig:dict_abl}
\vspace{-10pt}
\end{figure}

\textbf{Higher dictionary coverage leads to better performance.} We randomly mask out some of the entries in the dictionary at different probabilities.
We report the translation performance at different mask ratios in \autoref{fig:dict_abl}.
As the ratio of masked entries increases, the performance drops significantly.
This indicates that dictionaries that cover more words can lead to better performance.

\noindent \textbf{Dictionary with references to relevant words leads to better performance.}
When a corresponding entry is found for a word by \method, we would further explore other words referenced by this entry to provide more information.
We demonstrate that these links are more helpful by removing them.
As shown in \autoref{fig:dict_abl}, when these linked words are removed from dictionary entries of Manchu, the performance of \method drops no matter what the coverage of the dictionary is.

\subsection{How to make use of grammar knowledge.}

\textbf{Grammar book helps.}
\method without grammar book is not as good as \method with grammar. This is indicated by the performance boost from including grammar books. We perform a qualitative comparison between the outputs of \method with and without the short grammar on Manchu. Overall, the dictionary-only \method is able to capture the key content nouns in the sentences, but the full sentences generated are often not coherent or have the wrong sentence types (e.g. incorrectly generating a simple sentence instead of a question).
For example, \textit{``I have set out at the beginning of this month''} versus \textit{``I have set out on this new moon''}, \textit{``This one ordinary horse is said to be worth ten taels''} versus \textit{``This one horse ten ounces is said to be worth''}.
In the first example, \method with grammar better captures the meanings outside the surface meanings (``\textit{moon}'') of words and chooses the correct sense (``\textit{month}'') in cases of polysemous nouns.
In the second example, \method with grammar is able to translate the output in the correct order, even though the word order of Manchu (SOV) is different from that of English (SVO).

\noindent \textbf{Different chapters of the grammar book contribute differently.}
Most grammar books have chapters on different topics -- phonology, morphology, syntax, etc.
We experiment with including different chapters of the grammar book in the prompts and evaluate the outputs on Arapaho.
We find out that morphology chapters do not have significant improvements in the translation because the word stems and features are already identified by the morphological analysis.
For instance, the sentence \textit{``That's when this buffalo bull was used''} in Arapaho is \textit{``Ne'nih'ii'tonouneihit nehe' heneecee''}, the word stems and features are \textit{that-PAST.when-used-3.S this buffalo.bull}.  With the morphology chapter, the translation \textit{``This buffalo bull used (it).''} mistakenly consider ``buffalo'' as the subject, while without the chapters, the translation \textit{He used this buffalo bull} follows the order of the features and correctly identifies buffalo as the object.

\subsection{Comparing \method with a human baseline.}

To give the readers a better idea of \method's performance, we have asked one of the authors of this paper to mimic the behavior of \method in translating Manchu to English.
They have not been exposed to our Manchu data and have no prior knowledge of Manchu.
We ask them to finish the translation of 35 Manchu sentences in a limited amount of time (2 hours), which is longer than the time cost for LLMs.
We report both human and model's performance in \autoref{tab:hum_ba}.
The superior performance of human indicates that although LLMs can already learn a lot about a new language from its grammar book and dictionary, they are still not as good as human beings on this task.

\begin{table}
    \centering\small
    \begin{tabular}{cccc}
    \toprule
         & Few-Shot  & \method & Human \\
         \midrule
         BLEU&  0.82& 9.12  & 20.32   \\
         \bottomrule
    \end{tabular}
    \caption{Comparing the human baseline and \method on 35 Manchu sentences in translation.}
    \label{tab:hum_ba}
\end{table}

\subsection{\method as a benchmark for long-context understanding.}

\method really requires its backbone LLM to have good long-context understanding ability as grammar books can easily exceed 50000 tokens.
Not only does the LLM need a large context size, it must also be good at finding the relevant chapters and examples in the book to perform well in \method.
Therefore, we can use the performance with \method as a metric for comparing LLMs.
To demonstrate this possibility, we evaluate 4 backbone models on the mnc to eng task and report the results in \autoref{tab:bench}.
The performance gaps between different models are significant even for similar-sized models such as Olmo-7B and Mistral-7B.

\begin{table}
    \centering\small
    \begin{tabular}{cccc}
    \toprule
         Olmo-7B & Mistral-7B  & Mixtral-8x7B & GPT-4 \\
         \midrule
         0.54 &  3.71 & 4.4  & 10.8   \\
         \bottomrule
    \end{tabular}
    \caption{LingoLLM on top of different LLMs differentiates their long-context understanding ability.}
    \label{tab:bench}
\end{table}

\section{Lessons for Further Extending \method}

We ran into many obstacles and caveats while collecting linguistic descriptions and building \method.
Since some of these caveats are common to other endangered languages, we record them here for readers interested in extending \method or similar approaches to more languages.

\subsection{A large amount of linguistic descriptions are not easily tokenized.}
Many linguistic descriptions are scanned from physical books that involve typewritten and hand-written parts.
Converting them to plain text that can be tokenized is not easy, especially when there are non-Latin scripts or infrequent alphabets.
Things get even more complicated when there are complex hierarchies in the organization of grammar books or when the examples and grammar descriptions are interleaved in the book.
We had to give up adapting \method to several languages simply because of the digitization difficulty.

\subsection{Dictionaries do not have a universal interface.}
Interface for low-resource language' dictionaries vary from online dictionaries paired with different search algorithms, and digital PDFs to scanned books. The lack of a universal interface creates challenges in comparing the task performance across multiple low-resource languages as the interface used by one language is often not available in another language. The task performance for different low-resource languages is thus highly dependent on the implementation of the dictionaries. 

\subsection{Different types of linguistic descriptions often mismatch, creating a lot of trouble.}
Different types of linguistic descriptions for a low-resource language are often created separately by different authors with varying resources available, these mismatch causes confusion during translation. For example, the Gitksan morphological analyzer \citep{forbes-etal-2021-fst} is based on a different dictionary than the dictionary we use \citep{gitksan_mothertongues}. Some word stems such as \textit{jida} identified using the morphological analyzer cannot be found in the dictionary.

\section{Conclusion}

In this paper, we introduced \method, a novel approach for enabling LLMs to process endangered languages.
\method integrates linguistic descriptions such as grammar books and dictionaries, a critical resource that is often more available for endangered languages than extensive corpora.
\method has demonstrated remarkable improvements on multiple tasks across many languages.
Our work with \method highlights the potential of existing linguistic resources in the era of advanced LLMs and how they might make endangered languages more accessible in modern technological contexts.

\section*{Limitation}

We only experiment with 8 endangered and/or low-resource languages.
Due to the limited resources of native speakers of endangered languages, Our evaluation of the math reasoning, keyword-to-text, and word reordering is only on Manchu; we plan to extend to other languages.
We acknowledge the potential contamination in reasoning tasks because the original problems in high-resource languages are widely spread on the internet.
Lastly, \method has only been proven to work on languages with a Romanized script, which isn't something that every endangered language has.
We hope future works can build upon \method to better help the community of endangered languages.

\section*{Impact Statement}

Many endangered languages \textit{will} disappear within a few generations as their speakers die out.
The legends, myths, stories, songs, and other knowledge written in these languages will disappear with them.
While proper documentation can help preserve some aspects of these languages, \method sheds light on a more interactive way of preservation, adding a new tool to linguist's inventory.
In some sense, a model that can produce text in a language contains rich information about it.

Other than preservation, \method can also have a positive impact on current speakers of endangered languages, especially those who find it difficult to communicate in high-resource languages.
With the help of LLMs, they can have easier access to resources available in high-resource languages and have their voice heard by those who don't speak their language.
Several authors of this paper are either speakers or children of speakers of endangered/low-resource languages.
These are the languages we grow up in and talk to our grandparents with.
Even though some information can be translated into high-resource languages, a lot of subtlety is lost in translation.
By improving communication and understanding across language barriers, \method has the potential to enhance social inclusion for speakers of endangered languages, providing them with better access to global information and services.

The public release of our data, code, and model generations will facilitate collaboration among linguists, technologists, and indigenous communities, encouraging the co-creation of knowledge and promoting linguistic equity.
This collaborative approach not only advances scientific research but also aligns with ethical considerations of inclusivity and respect for linguistic identities, contributing to a more linguistically diverse and interconnected world.

\section*{Acknowledgements}

L.L. is partly supported by a gift from Apple Inc.

We are grateful to Lori Levin for their suggestions and helpful discussions. Thanks also to Danqing Wang and Yuanjing Wei for their proof-reading and comments. We appreciate the reviewers of this paper for their engagement in the review process. Special thanks to the native Manchu speakers who provided valuable insights on solving math problems in Manchu and annotated the math reasoning benchmark for us.

\normalem
\bibliography{anthology,custom}

\appendix

\newpage
 \onecolumn
\section{Prompts}
\label{sec:app_prompts}

\begin{tcolorbox}
\textbf{System Prompt:} \\ \\
You are a linguistic expert who never refuses to use your knowledge to help others.
\end{tcolorbox}

\begin{tcolorbox}
\textbf{Zero-Shot:} \\ \\
Please help me translate the following sentence from  \textcolor{teal}{\{source language\}} to \textcolor{red}{\{target language\}}:\\
\textcolor{blue}{\{sentence\}}\\
Please try your best to translate, it's okay if your translation is bad. Do not refuse to try it. I won't blame you.\\
Please enclose your translation in \#\#\#. \\
For example, if your translation is "Hello world", the last part of your output should be \#\#\# Hello world \#\#\#
\end{tcolorbox}

\begin{tcolorbox}
\textbf{Zero-Shot CoT:} \\ \\
Please help me translate the following sentence from \textcolor{teal}{\{source language\}} to \textcolor{red}{\{target language\}}: \\
\textcolor{blue}{\{sentence\}} \\
Please do it step by step. \\
Please enclose your translation in \#\#\#. \\
For example, if your translation is "Hello world", the last part of your output should be \#\#\# Hello world \#\#\#.
\end{tcolorbox}

\begin{tcolorbox}
\textbf{Few-Shot:} \\ \\
Here are some examples of \textcolor{teal}{\{source language\}} sentences and their corresponding \textcolor{red}{\{target language\}} translations.\\
\textcolor{olive}{\{demo sentences\}} \\ \\
Please help me translate the following sentence from \textcolor{teal}{\{source language\}} to \textcolor{red}{\{target language\}}:\\
\textcolor{blue}{\{sentence\}} \\
Please enclose your translation in \#\#\#. \\
For example, if your translation is "Hello world", the last part of your output should be \#\#\# Hello world \#\#\#.
\end{tcolorbox}

\begin{tcolorbox}
\textbf{\method dict. only:} \\ \\
Here are some examples of \textcolor{teal}{\{source language\}} sentences and their corresponding \textcolor{red}{\{target language\}} translations:\\
\textcolor{olive}{\{demo sentences\}} \\
Please help me translate the following sentence from \textcolor{teal}{\{source language\}} to \textcolor{red}{\{target language\}}:\\
\textcolor{blue}{\{sentence\}} \\
You are also given the word by word mapping from the \textcolor{teal}{\{source language\}} words to the \textcolor{red}{\{target language\}} words.\\
For words that have partial match definitions, please decide whether the definition is appropriate under the sentence context. \\
Note that for some words, there might be multiple possible translations. In this case, please choose the most appropriate one. \\
Note that for some words, they might be derived from a more basic form, we call this the parent word. The parents are also given in the word by word translation. \\
Here is the dictionary entry for each individual word in the source sentence:\\
\textcolor{magenta}{\{wordbyword mapping\}} \\
Please first explain what each word means in \textcolor{red}{\{target language\}}  and then translate.\\
Remember your source sentence is: \\
\textcolor{blue}{\{sentence\}} \\
Please enclose your translation in \#\#\#. \\
For example, if your translation is "Hello world", the last part of your output should be \#\#\#Hello world\#\#\#.
\end{tcolorbox}

\begin{tcolorbox}
\textbf{\method:} \\ \\
You are given this \textcolor{teal}{\{source langauge\}} grammar book. Feel free to rely on the grammar rules in the book in your translation.\\
\textcolor{orange}{\{grammar\}}
Please help me translate the following sentence from \textcolor{teal}{\{source langauge\}} to \textcolor{red}{\{target language\}}:\\
\textcolor{blue}{\{sentence\}} \\
You are also given the word by word mapping from the \textcolor{teal}{\{source langauge\}} words to the \textcolor{red}{\{target language\}} words.\\
Note that for some words, there might be multiple possible translations. In this case, please choose the most appropriate one.\\
Note that for some words, they might be derived from a more basic form, we call this the parent word. The parents are also given in the word by word translation.\\
\textcolor{magenta}{\{wordbyword mapping\}} \\
Given the above book and word for word mapping.
Please first annotate the meaning and grammatical features of each word in the sentence according to their suffixes and the grammar book.\\
For each noun, please annotate its number and case.\\
For each verb, please annotate its tense.\\
For each verb, please annotate its voice.\\
For each verb, please annotate its form.\\
Please figure out what the subject and object of each verb is.\\
After annotation, please translate the sentence into \textcolor{red}{\{target language\}}  and enclose your translation in \#\#\#.
\end{tcolorbox}

\section{Linguistic Descriptions} \label{app:lind}

\begin{table*}[t!]\small\centering
\begin{tabular}{llll}
\toprule
Language                & Manchu & Gitksan & Arapaho       \\
\midrule
Dictionary  & \citet{norman2020comprehensive} & \citet{gitksan_mothertongues} & \citet{kazeminejad-etal-2017-arapaho-dictionary} \\
Grammar     & \citet{gorelova2002manchu} & \citet{rigsby1986gitksan}  &\citet{c556855e-04da-352f-8f93-78660f9dc948-arapaho-grammar}      \\
Morphological Analyzer & \citet{buleku2023} & \citet{forbes-etal-2021-fst} & \citet{moeller-etal-2018-neural}
\\ \midrule
Language                & BriBri & Tsez & Wolof       \\
\midrule
Dictionary & \citet{bribri_dict} & \citet{ginn-etal-2023-findings}& \citet{wolof_dict}\\
Grammar & \citet{brirbri_gramática} &  N/A& \citet{wolof_grammar}\\
Morphological Analyzer & \citet{bribri_fst}& \citet{ginn-etal-2023-findings}&
N/A\\ \midrule
Language                & Uspenteko & Natugu &        \\
\midrule
Dictionary & \citet{ginn-etal-2023-findings}& \citet{ginn-etal-2023-findings}& \\
Grammar & N/A& \citet{Alfarano2021Grammaire}& \\
Morphological Analyzer & \citet{ginn-etal-2023-findings}& \citet{ginn-etal-2023-findings}&
\\ \bottomrule
\end{tabular}
\caption{Linguistic descriptions we use for different endangered languages.}
\label{tab:resources}
\end{table*}

\section{Language and Their ISO 639-3 Code} \label{app:code}

\begin{itemize}
    \item mnc - Manchu.
    \item git - Gitksan.
    \item usp - Uspanteko.
    \item ntu - Natugu.
    \item ddo - Tsez.
    \item wol - Wolof.
    \item arp - Arapaho.
    \item bzd - Bribri.
\end{itemize}

\newpage

\section{More translation examples}

\begin{table*}[th]
\small
\centering
\begin{tabular}{p{6cm}p{6cm}p{1.4cm}p{1cm}}
\toprule
                \textbf{Ground Truth}& \method& 
                \textbf{BLEURT}& \textbf{Rank} \\ \midrule
 Now where are you going?& Where are you going now?	& 0.86&1\%\\ \midrule
Among your many classmates, how many are Chinese and how many are Korean?	&  How many Chinese people and how many Koreans are there among your numerous students?	&  0.74&    5\%\\ \midrule
A letter from home is worth ten thousand liang of gold.	&  The letter of the house is worth ten thousand ounces of gold.	&  0.69&      11\%\\ \midrule
I'm going to Beijing (the capital city)	&  I am going toward the city's capital.	& 0.65&     20\%\\
\midrule
What books does he explain?& What book are we explaining?	& 0.62	& 30\%
\\
\midrule
Don't worry for us, it's no big deal.	& You need not be very distressed on our account, it doesn't matter.	& 0.59& 40\%\\
\midrule
I live in Liaodong city.&  I have resided in the inner part of the walled city of Liaodong.	& 0.53	&50\%
\\
\midrule
Master, light a lamp and bring it&  Having lit the lamp, bring it (along), master.	& 0.50	&60\%\\
\midrule
I'm a Korean person, I don't walk with familiarity in places of China	& I, even if a person of Korea, was not sad about walking the land of the Chinese.	& 0.47	& 70\%\\
\midrule
Why do you learn Chinese language?	& You, teach that Chinese person's book, how?	& 0.44& 80\%\\
\midrule
This is very well, we should go together!	& If so, we will likely do it together, the partridge.	& 0.40& 90\%\\
\bottomrule
\end{tabular}
\caption{Example translations produced by \method, compared to ground truth translation and the few-shot baseline.
Note that the translations from few-shot prompting are nonsensical and completely irrelevant to the actual translation.
More examples in Appendix}
\label{tab:translate_ex_app}
\end{table*}

\newpage

\section{Math reasoning examples}

\paragraph{Question in Manchu:}
Balder hvqin deri muke tunggiyeme gaimbi. erde sunja hunio, yamji ninggun hunio tunggiyembi. emu hunio de sunja moro biqi, tere inenggi moro udu muke bahambi?

\paragraph{Question in English (Ground Truth):}
Baldur gets water from a well. He gets 5 pails of water every morning and 6 pails of water every afternoon. If each pail contains 5 liters of water, how many liters of water does he get every day?

\paragraph{Question in English (\method):}

Balder, early in the morning, picks up water from the well. He takes five buckets in the evening, and six buckets in the morning. If one bucket equals five bowls, how many bowls of water does he get in a day?

\noindent\textbf{Verbatim Output of \method:}

\begin{tcolorbox}
\small
\textbf{Let's break down the sentence and annotate each word as per the grammar rules provided:}

Balder: {key} is a person's given name: {key} - Proper noun, Nominative case, singular\\
hvqin: a well - Noun, Nominative case, singular\\
deri: (ablative particle) from, than, via, by way of - Ablative case marker\\
muke: 1. water 2. river, stream - Noun, Nominative case, singular\\
tunggiyeme: 1.Pick up - Verb, Affirmative, Subjective voice, Present/Future tense\\
gaimbi.: 1. to take, to take away, to take off - Verb, Affirmative, Subjective voice, Present/Future tense\\
erde: early, early in the morning - Adverb\\
sunja: five - Numeral, used as an adverb here\\
hunio: water bucket, pail, tub - Noun, Nominative case, singular\\
yamji: evening - Adverb\\
ninggun: six - Numeral, used as an adverb here\\
hunio: water bucket, pail, tub - Noun, Nominative case, singular\\
tunggiyembi: to pick up, to gather up - Verb, Affirmative, Subjective voice, Present/Future tense\\
emu: one - Numeral, used as an adverb here\\
hunio: water bucket, pail, tub - Noun, Dative/Locative case, singular (indicated by the following particle 'de')\\
de: (dative/locative particle) to, in, at, on, by - Dative/Locative case marker\\
sunja: five - Numeral, used as an adverb here\\
moro: bowl, a dry quart - Noun, Nominative case, singular\\
biqi: conditional converb form of 'bimbi' - to exist, to be - Converb, Conditional\\
tere: that - Demonstrative pronoun, Nominative case, singular\\
inenggi: day, daytime - Noun, Nominative case, singular\\
moro: bowl, a dry quart - Noun, Nominative case, singular\\
udu: "How many?", "How much?" - Interrogative pronoun\\
muke: water - Noun, Accusative case, singular (indicated by the verb 'bahambi' which requires an object in the accusative case)\\
bahambi?: to get, to obtain - Verb, Interrogative form, Subjective voice, Present/Future tense\\

\textbf{Subject and object identification:}\\
- Balder is the subject of the verb 'tunggiyeme' (to pick up).\\
- 'Muke' (water) is the object of the verb 'tunggiyeme' (to pick up).\\
- 'Hunio' (water bucket) is the object of the verb 'gaimbi' (to take).\\
- 'Muke' (water) is the object of the verb 'bahambi' (to get).\\

\textbf{Now let's translate the sentence into English:}

"Balder, early in the morning, picks up water from the well. He takes five buckets in the evening, and six buckets in the morning. If one bucket equals five bowls, how many bowls of water does he get in a day?"
\\
\textbf{Now, let's solve the math question:}

In the evening, Balder takes 5 buckets.
In the morning, he takes 6 buckets.
One bucket equals 5 bowls.

Total buckets in a day = 5 (evening) + 6 (morning) = 11 buckets
Total bowls in a day = 11 buckets * 5 bowls per bucket = 55 bowls

So, the answer is: 55
\end{tcolorbox}

\newpage

\noindent\textbf{Question in Manchu:} gvsin ninggun (36) penguin ni nimanggi tala de xun fiyakamahabi. baji ome ilan i emken (1/3) mederi qi dosifi ebxeme genefi, jai ilan i emken (1/3) dunggu de jaka jeme genefi. geli ya labdu penguin kemuni xun fiyakame ilihabi?

\noindent\textbf{Question in English (verbatim):} There are 36 penguins sunbathing in the snow. One-third of them jump in and swim in the ocean. Another one-third go inside the cave to eat their dinner. How many penguins are still left sunbathing?

\noindent\textbf{Question in English (\method):} Thirty-six penguins on the snowy plain sled every day. A little more than one-third of them hurriedly entered the sea, and another one-third went into the cave to eat something. How many penguins are still sledding every day?

\begin{tcolorbox}
\scriptsize
\textbf{Let's begin by annotating the sentence according to the grammar rules provided:}\\
\\
gvsin ninggun (36) penguin ni nimanggi tala de xun fiyakamahabi.\\
- gvsin: thirty [indefinite number]\\
- ninggun: six [indefinite number]\\
- (36): thirty-six [composite number formed by gvsin + ninggun]\\
- penguin: penguin [Nominative, singular]\\
- ni: (genitive particle) [marks the following noun phrase as genitive]\\
- nimanggi: snow [Nominative, singular]\\
- tala: plain, steppe [Dative/Locative, singular]\\
- de: (dative/locative particle) [marks the location "on the plain"]\\
- xun: day [Nominative, singular]\\
- fiyakamahabi: to sled (perfective aspect, past tense, affirmative, subjective voice)\\
\\
baji ome ilan i emken (1/3) mederi qi dosifi ebxeme genefi,\\
- baji: a little bit more [adverb]\\
- ome: to become (imperfect converb) [marks an ongoing action]\\
- ilan: three [indefinite number]\\
- i: (genitive particle) [marks the following noun phrase as genitive]\\
- emken: one-third [fraction formed by ilan + i + emken]\\
- (1/3): one-third [fraction]\\
- mederi: sea [Ablative, singular]\\
- qi: than [comparative particle]\\
- dosifi: to enter (perfect converb) [marks a completed action]\\
- ebxeme: hurriedly [adverb]\\
- genefi: to go (perfect converb) [marks a completed action]\\
\\
jai ilan i emken (1/3) dunggu de jaka jeme genefi.\\
- jai: and [conjunction]\\
- ilan: three [indefinite number]\\
- i: (genitive particle) [marks the following noun phrase as genitive]\\
- emken: one-third [fraction formed by ilan + i + emken]\\
- (1/3): one-third [fraction]\\
- dunggu: cave [Dative/Locative, singular]\\
- de: (dative/locative particle) [marks the location "in the cave"]\\
- jaka: thing, object [Nominative, singular]\\
- jeme: to eat (imperfect converb) [marks an ongoing action]\\
- genefi: to go (perfect converb) [marks a completed action]\\
\\
geli ya labdu penguin kemuni xun fiyakame ilihabi?\\
- geli: still [adverb]\\
- ya: which [interrogative pronoun]\\
- labdu: many [indefinite number]\\
- penguin: penguin [Nominative, singular]\\
- kemuni: often [adverb]\\
- xun: day [Nominative, singular]\\
- fiyakame: to sled (imperfect converb) [marks an ongoing action]\\
- ilihabi: to stand (interrogative form) [marks a question]\\
\\
\textbf{Now, let's translate the sentence into English:}\\
\\
Thirty-six penguins on the snowy plain sled every day. A little more than one-third of them hurriedly entered the sea, and another one-third went into the cave to eat something. How many penguins are still sledding every day?\\
\\
For the math question, we need to find out how many penguins are left sledding every day after subtracting the two groups of one-third each:\\
\\
There are 36 penguins in total.\\
One-third of 36 is 12 (because 36 / 3 = 12).\\
Two groups of one-third would be 24 (because 12 * 2 = 24).\\
Subtracting these two groups from the total gives us 36 - 24 = 12.\\
\\
So, the answer is: 12.
\end{tcolorbox}

\end{document}